\documentclass[letterpaper, 10 pt, conference]{ieeeconf}  % Comment this line out if you need a4paper

\IEEEoverridecommandlockouts                              % This command is only needed if 
\usepackage{graphicx}
\usepackage{multirow} % kmk
\usepackage{amssymb}  % assumes amsmath package installed
\usepackage{subcaption}

\usepackage{amsmath} % syh
\usepackage{amssymb} % syh-option
\usepackage{xcolor}
\usepackage{url}

\newcommand{\BL}[1]{}

\title{\LARGE \bf
A Shared-Autonomy Construction Robotic System for Overhead Works 
}
\author{David Minkwan Kim$^{*1}$, 
K. M. Brian Lee$^{*2}$, 
Yong Hyeok Seo$^{*1}$, 
Nikola Raicevic$^{*2}$, 
Runfa Blark Li$^{*2}$, \\
Kehan Long$^{2}$, 
Chan Seon Yoon$^{3}$, 
Dong Min Kang$^{3}$, 
Byeong Jo Lim$^{3}$, \\
Young Pyoung Kim$^{3}$, 
Nikolay Atanasov$^{2}$, 
Truong Nguyen$^{2}$,
Se Woong Jun$^{1}$,  
Young Ouk Kim$^{1}$%
\thanks{This work was supported by the Ministry of Trade, Industry and Energy (MOTIE), Korea, under the Strategic Technology Development Program, supervised by the Korea Institute for Advancement of Technology (KIAT) [Grant No. P0026052]. *These authors contributed equally to this work.}% <- this % stops a space
\thanks{$^{1}$Intelligent Robotics Research Center, Korea Electronics Technology Institute (KETI), Seongnam, Korea.
        {\tt\scriptsize \{kmk9846, syh4661, daniel, kimyo\}@keti.re.kr}}%
\thanks{$^{2}$ Dept. of Electrical and Computer Engineering, University of California, San Diego (UCSD), La Jolla, CA, USA.
        {\tt\scriptsize \{kmblee, 
        nmarinoraitsevits, runfa, k3long, natanasov, tqn001\}@ucsd.edu}}%
\thanks{$^{3}$ITONE Inc., Seoul, Korea. 
        {\tt\scriptsize \{parcae, kangdm, limbj, mrkims\}@it-1.kr}}%
}

\usepackage[hidelinks]{hyperref} % clickable URLs (arXiv v1.5 does not auto-load hyperref)
\begin{document}

\maketitle
\thispagestyle{empty}
\pagestyle{empty}

%%%%%%%%%%%%%%%%%%%%%%%%%%%%%%%%%%%%%%%%%%%%%%%%%%%%%%%%%%%%%%%%%%%%%%%%%%%%%%%%
\begin{abstract}
% A promising solution to labor shortages in construction is to improve job safety and comfort using robotic systems.
We present the ongoing development of a robotic system for overhead work such as ceiling drilling.
% \NA{Make it more precise what "ceiling works" means. It may be good to start with an overview sentence about construction robotics.}. 
The hardware platform comprises a mobile base with a two-stage lift, on which a bimanual torso is mounted with a custom-designed drilling end effector and RGB-D cameras. 
To support teleoperation in dynamic environments with limited visibility, we use Gaussian splatting for online 3D reconstruction and introduce motion parameters to model moving objects.
% \NA{Instead of listing an algorithm that no one knows, I suggest describing the main idea, e.g., "..we use Gaussian Splatting and introduce motion parameters for each Gaussian to enable reconstruction of dynamic environments." } 
For safe operation around dynamic obstacles, we developed a neural configuration-space barrier approach for planning and control. 
% \NA{Split this in a separate sentence. Remove the reference. Explain what the objective is before mentioning an approach.}  
Initial feasibility studies demonstrate the capability of the hardware in drilling, bolting, and anchoring, and the software in safe teleoperation in a dynamic environment. 
Experiment videos and results are available at \url{https://kmk9846.github.io/Construction-Robot/}.

\end{abstract}

\section{INTRODUCTION}
% syh -draft 과제 연차보고서 내용 일부 발췌 및 문장 적용

%건설현장 내 안전관련 이슈로 노동기피현상으로 인한 사업장 내 인력난이 대두되고 있다. 이에 본 연구에서는 이를 위한 솔루션으로서 로봇을 활용한 현장 건설 작업자의 노하우를 로봇운영에 활용하되 비직관적인 로봇 원격제어에 의한 부담 및 작업 피로도 유발을 인공지능 가이던스 기술을 활용하여 극복하고자 한다. 이에 초창기 모델로 고소작업이 가능한 하드웨어를 구상하고, 이와 더불어 기존에 가시권에서 조작자의 육안으로 원격제어를 하는 형태에서 비가시권 원격지 작업의 비효율성을 타개하고자 거리감 및 임장감 확대를 위해 고수주느이 인공지능 기반 3D 기술을 적용하여, 효율적인 원격제어 건설로봇 시스템을 구성하는 데에 초도연구를 진행하였다.
Construction faces worsening labor shortages due to the tough and risky nature of the job~\cite{constructiondive2018}, often referred to as “dirty, dull, and dangerous (DDD)” tasks. 
This perception contributes to an aging workforce and, in turn, a higher risk of injuries~\cite{Schwatka2012}.
A growing number of commercial systems, such as the Hilti Jaibot~\cite{hilti_jaibot}
% \footnote{\scriptsize \url{https://www.hilti.com/content/hilti/W1/US/en/business/business/trends/jaibot.html}} 
and CSC Robo Drillcorpio~\cite{csc_drillcorpio},
% \footnote{\scriptsize \url{https://cscrobotic.com/en/drillcorpio-model-d3/}}\NA{Generally footnote urls are discouraged. It may be better to use a regular citation.}
are being developed to assist construction work.
We aim to advance this effort by deploying state-of-the-art techniques from mobile and field robotics to practical systems that improve job safety and comfort.

% While commercial systems such as the Hilti Jaibot~\cite{hilti_jaibot}\BL{Replace with footnotes} and CSC ROBO's Drillcorpio~\cite{csc_drillcorpio}\BL{Replace with footnotes} demonstrate the growing industrial adoption of ceiling construction automation, our system further integrates teleoperation, dynamic 3D reconstruction, and real-time planning and control for safer operation in complex and dynamic environments. We aim to develop robotic systems that improve safety and reduce physical strain, making such jobs safer and more feasible.

% 1. 기존 로봇/연구와 비교 
\begin{figure}[t]
    \centering
    \includegraphics[width=0.95\linewidth]{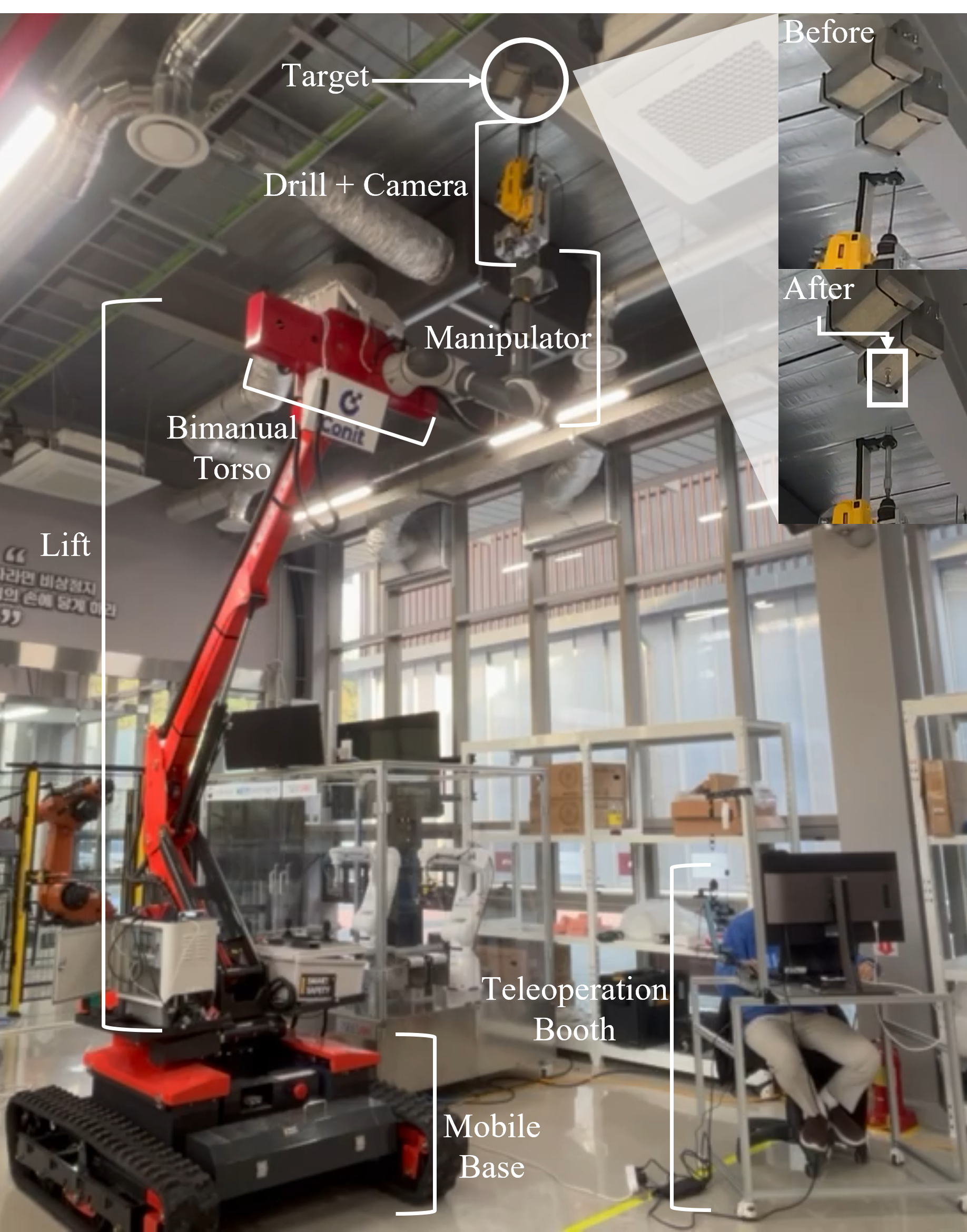}
    \caption{Our system performing drilling, bolting, and anchoring tasks in a laboratory setting. Video available\protect \footnotemark.}
    \label{fig:robot_whole_hardware}
    \vspace{-2ex}
\end{figure}

% 3. 기존 로봇/연구와 비교 

% \begin{figure*}[!t]
%     \centering
%     \includegraphics[scale=0.45]{ieeeconf/1.Hardware/figs/robot_whole_hardware_2.png}
%     % \includegraphics[scale=0.6]{ieeeconf/1.Hardware/figs/robot_real_image.png}
%     \caption {This figure illustrates the overall system of the construction robot. The robot mainly consists of a mobile base and a manipulator. The operator can control the robot from a teleoperation booth, enabling operation in a BVLOS~\cite{drones2020013} environment.}
%     \label{fig:robot_whole_hardware}
% \end{figure*}

This paper introduces the ongoing development of a robotic system designed for overhead ceiling work, one of the most physically straining jobs in construction. 
We developed a mobile lift system (Fig.~\ref{fig:robot_whole_hardware}) equipped with a torso that carries up to two manipulators, each with a drilling end effector and sensing payloads to perform overhead tasks such as drilling on ceilings or attaching light fixtures.
% \NA{Give examples of the types of tasks that we may be doing, e.g., drilling, attaching light fixtures, etc.?}
A teleoperation system is designed for intuitive operation, to best utilize the expertise of experienced on-site workers in robot operation.

A key challenge is ensuring safety and situational awareness when teleoperating in real construction sites with many moving objects and occlusions.
For situational awareness, we use our prior work on DynaGSLAM~\cite{dynagslam} to build a 3D Gaussian splat (3DGS) reconstruction of dynamic environments to allow novel-view synthesis for beyond-visual-line-of-sight (BVLOS) teleoperation as is common in aerial robots~\cite{drones2020013}.
% \NA{Split this into two sentences and explain first what the main problems are before offering solutions. 
% Then, you can emphasize that (1) these are our works and (2) they contribute to a particular aspect of the problem in a novel way.} 
For safe operation in dynamic environments, our prior work in neural configuration-space barriers (NCSB)~\cite{neural_cspace_barrier} is used to plan joint trajectories to desired end-effector poses, and to `filter' both teleoperation and autonomy commands for collision avoidance against dynamic obstacles. 

\footnotetext{\scriptsize Video: \href{https://www.kmblee.dev/videos/construction_workshop}{\nolinkurl{www.kmblee.dev/videos/construction_workshop}}}
We present an initial feasibility study with a single-arm prototype that shows the capability of the hardware platform to drill, anchor and bolt on ceilings.
Further experimental results demonstrate safe planning and control in dynamic environments for collision avoidance using NCSB, and 3D reconstruction using DynaGSLAM.
% 2. 기존 로봇/연구와 비교 
We view the main contribution of this paper as the initial demonstration of state-of-the-art robotics techniques applied in a practical construction robotic system.
% We hope that our efforts will enhance the appeal of construction as a career, such as allowing working from home as do office workers.

% that will eventually allow construction workers to work from home as they desire.

% To address this challenge, this study explores the use of robotics to  
% To mitigate the burden and fatigue associated with unintuitive remote control, AI-guided assistance technologies are incorporated. 
% An initial prototype capable of performing ceiling construction tasks has been developed. 
% Furthermore, to overcome the limitations of conventional line-of-sight-based teleoperation, advanced AI-driven 3D perception technologies are employed to enhance depth perception and situational awareness in Beyond Visual Line of Sight (BVLOS)~\cite{drones2020013} environments. 
% This research represents a preliminary effort to develop an efficient remote-controlled construction robot for ceiling construction tasks. % 영어 번연 kmk, syh 수정정

% Related Work 대신에 Intro 안에서 기존에 존재하는 시스템 + 어떻게 다른지 설명 

% "AI-driven 3D perception 내용 인트로는 일단 간략하게만 넣었습니다. UCSD 측에서 develop 해주시면 감사드립니다.

% \section{RELATED WORKS} % 넣을지 말지 고민중입니다. syh
% Deillcorpio
% HILTI jayrobot

\section{HARDWARE PLATFORM} %Hardware vs Hardware platform

We designed and built a robotic system and a teleoperation booth as shown in Fig.~\ref{fig:robot_whole_hardware} to enable BVLOS teleoperation for construction tasks. We briefly describe the system.
% To enable construction tasks in Beyond Visual Line of Sight (BVLOS)~\cite{drones2020013} environments, we designed the system shown in Fig.~\ref{fig:robot_whole_hardware}. 
% The system comprises three main components: \textbf{Lift and Manipulator}, \textbf{Mobile Base}, and \textbf{Teleoperation Booth}.

\subsection{Robotic System}
%     - Ver. 1, RB-10 ⇒ Ver. 2 Doosan(in future work)
%     - Torque specification
%     - Mounting design
%     - End-effector
%     - Sensors

\begin{figure}[!t]
    \centering
    \includegraphics[width=\linewidth]{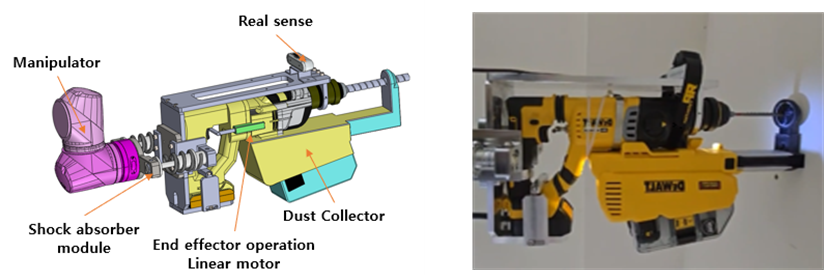}
    \caption{A custom drilling end-effector assembly.}
    \label{fig:robot_damping_system}
    \vspace{-2ex}
\end{figure}

The robot hardware consists of a mobile base, a two-stage lift, and a manipulator. The two-stage lift enables vertical motion of a torso from 437~mm to 3,680~mm. The torso is designed to accept up to two manipulators. Currently, a Rainbow Robotics RB10-1300 manipulator is horizontally mounted on the torso and supports up to 10~kg of payload. A custom end-effector (Fig.~\ref{fig:robot_damping_system}) is attached for drilling, bolting, and anchoring, with a damping module placed between the drill and manipulator to reduce load impact. To support teleoperation, a pan-tilt VR camera on top of the torso provides real-time 3D visualization, while a RealSense D435 camera above the drill offers local visual feedback during tasks.
The mobile base measures 2,660~mm × 1,350~mm × 1,710~mm and reaches speeds up to 2.4~km/h including the lift and manipulator. Both the base and lift can be remotely controlled via UART communication. The system is optimized for ultra-low-latency control.

% A drill, designed for ceiling drilling, bolting, anchoring, is installed at the end-effector of the manipulator. 
% To reduce the stress imposed on the manipulator during construction tasks, a damping system (illustrated in Fig.~\ref{fig:robot_damping_system}) is integrated between the drill and the manipulator to absorb shock loads.

% Damping system :
% % damping system fig (syh)
% \begin{itemize}
%   \item \textbf{Spring Specification}: \\
%   \quad 750~\text{LBS/in}, \quad 13.39~\text{kg/mm} \\
%   \quad Equivalent to $1.31 \times 10^8~\text{N/m}$
  
%   \item \textbf{Total Mass (including hammer drill, bracket, etc.)}: \\
%   \quad $10~\text{kg}$
  
%   \item \textbf{Natural Frequency}: \\
%   \quad $114.61~\text{rad/s} \quad (18.24~\text{Hz})$ \\
%   \quad Compared to the excitation frequency of $50~\text{Hz}$, the risk of resonance is low. \\
%   \quad \textit{Note: Resonance typically occurs within} $\pm5\%$ \textit{of the system's natural frequency.}
  
%   \item \textbf{Damping Ratio}: \\
%   \quad $\zeta = 0.436$ \\
%   \quad Since $\zeta < 1$, the system is underdamped but within an appropriate range.
% \end{itemize}

% \subsection{Mobile Base}

% \BL{We could delete Figs 3 and 4.}

% \begin{figure}[!t]
%     \centering
%     \includegraphics[scale=0.3]{ieeeconf/1.Hardware/figs/control_interface.png}
%     \caption{The control interface for operating the mobile base and lift.}
%     \label{fig:control_interface}
% \end{figure}

%     - Hardware is off-the-shelf, software
%     - TCP control software

\subsection{Teleoperation Booth}
% As shown in Figure~\ref{fig:teleoperation_booth},
% We developed a teleoperation booth for controlling the construction robot in BVLOS~\cite{drones2020013}. 
The teleoperation booth is equipped with an operator PC, a monitor for visual inspection of the site, as well as a haptic device for control.
The control commands are transmitted from the operator PC to the robot PC via WebRTC~\cite{Tiberkak2023WebRTC} over a 5G network, enabling low-latency operation even in BVLOS settings.
For intuitive manipulation, the monitor offers photorealistic visual feedback of the site using DynaGSLAM~\cite{dynagslam} (Sec.~\ref{sec:dynagslam}), as well as
physical feedback and commands through a haptic device based on GELLO~\cite{wu2023gello} (Sec.~\ref{sec:teleop}).

% To assist with intuitive manipulation and ensure safe robot operation, the teleoperation commands are ``filtered'' with a control barrier function (CBF)~(Sec.~\ref{sec:cbf}).
% The teleoperation booth enables remote BVLOS operation via a monitor and haptic device for intuitive robot control. A control barrier function (CBF) filter ensures safe operation (Fig.~\ref{fig:overview_diagram}).

%Block Diagram
\begin{figure}[!t]
    \centering
    \includegraphics[width=\linewidth]{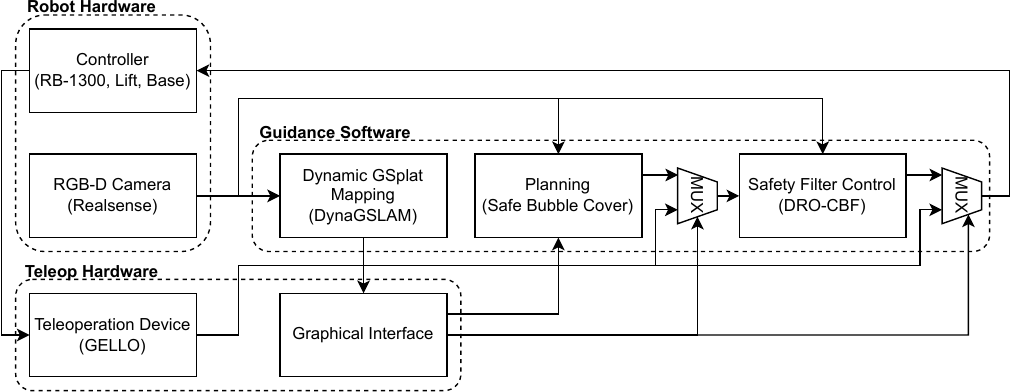}
    \caption{System overview}
    \label{fig:overview_diagram}
    \vspace{-2ex}
\end{figure}

% \begin{figure}[!t]
%     \centering
%     \includegraphics[scale=0.25]{ieeeconf/1.Hardware/figs/teleoperationbooth1.png}
%     \includegraphics[scale=0.25]{ieeeconf/1.Hardware/figs/teleoperationbooth2.png}
%     \caption{Teleoperation booth interface.}
%     \label{fig:teleoperation_booth}
% \end{figure}
%     - Design figure
%     - Haptic Sensing
% - Specification Table (IT-ONE) %kmk

\section{SOFTWARE}

An overview of the system is shown in Fig.~\ref{fig:overview_diagram}.
Given RGB-D images from the Realsense D435 camera, we construct a 3D Gaussian splat for visual teleoperation using DynaGSLAM \cite{dynagslam}, which is used for visual feedback in operator graphical interface.
An NCSB \cite{neural_cspace_barrier} is constructed from the RGB-D images for safe planning and control. 
% The planning and control modules uses the safe bubble cover~\cite{safe_bubble_cover,long24rss} algorithm to generate a joint trajectory from the current configuration to a drilling location specified by the worker, and a distributionally robust CBF (D-RO CBF) for safe execution in a dynamic environment~\cite{long24rss,kehan_dro_cbf}.
% The planne trajectory is executed using DR-CBF 

\subsection{Dynamic Gaussian Splat Mapping}\label{sec:dynagslam}
We use DynaGSLAM~\cite{dynagslam} from our prior work to provide photorealistic visual feedback of dynamic, cluttered construction sites.
Given RGB-D images of a dynamic scene from the Realsense D-435 camera, DynaGSLAM reconstructs the scene as a set of Gaussian blobs defined as $\mathcal{G} = \{ (\mathbf{m}_{i}(t), \Sigma_i,  \alpha_{i}, \mathbf{sh}_{i}) \}$, with means $\mathbf{m}_i(t)$, covariances $\Sigma_i$, transparency $\alpha_{i}$, and spherical harmonics coefficients $\mathbf{sh}_{i}$.
Importantly, DynaGSLAM produces photorealistic rendering of moving objects by formulating the means $\mathbf{m}_{i}(t)$ as time-varying and modeling their temporal evolution using cubic Hermite splines. For computational efficiency, the interpolation parameters of $\mathbf{m}_{i}(t)$ are directly initialized using a novel management strategy \cite[Sec. 5.2]{dynagslam} that fuses previous interpolation parameters with the current optical flow and RGB-D pointcloud.
The Gaussians are rendered and optimized against the RGB-D images using alpha blending:
\begin{equation}
    I(\mathbf{v}, t) = \sum_{g_{i} \in \mathcal{G}(t)} c_{i} f(\mathbf{v}, g_{i}) \prod_{i} (1 - f(\mathbf{v}, g_{i}) ),
\end{equation}
where $c_{i}$ is the color of Gaussian $g_i$ given spherical harmonics $\mathbf{sh}_{i}$, and $f(\mathbf{v}, g_{i}) =  \alpha_i N_{\text{2D}}(\mathbf{v}; P\mathbf{m}_i(t), P \Sigma_i P^{T})$ is the contribution of Gaussian $g_{i}$ at pixel $\mathbf{v}$ and time $t$. $P$ is the affine projection transform to the image plane, and the Gaussians are ordered relative to the query viewpoint.

%Autonomy
\subsection{Neural Configuration-Space Barrier}\label{sec:cbf}
% \NA{Describe the problem here and explain that NCSB is our work. For example, "Given a scene reconstruction in the form of Gaussian Splats, we consider the problem of planning and executing the motion of a robot arm to achieve a desired end-effector configuration subject to safety constraints." Then, explain how we formulate the safety constraints with respect to the environment reconstruction and the robot body.}
Given an RGB-D pointcloud $\mathcal{P}$, we consider the problem of planning and executing a manipulator trajectory that achieves a desired end-effector pose while avoiding dynamic obstacles.
This safety requirement is encoded as an NCSB, which is defined using the configuration-space distance field (CDF)~\cite{li2024cdf_rss}.
Let $\mathbf{q} \in \mathbb{R}^{N}$ be the manipulator's configuration, and $\mathcal{B}(\mathbf{q}) \subset \mathbb{R}^{3}$ the manipulator's geometry in the workspace. The CDF $d: \mathbb{R}^{N} \times \mathbb{R}^{3} \rightarrow \mathbb{R}$ is the closest distance in the configuration space (i.e. joint angles) to collision with a workspace point~\cite{li2024cdf_rss}:
\begin{equation}
    d(\mathbf{q}, \mathbf{p}) = \min_{\mathbf{q}^\star} ||\mathbf{q} - \mathbf{q}^\star||,\text{ s.t. }\mathbf{p} \in \partial\mathcal{B}(\mathbf{q}^\star).
\end{equation}
The CDF is approximated with a neural model $\hat{d}_{\theta} \approx d$ with weights $\theta$. Given the pointcloud $\mathcal{P}$, the NCSB is $h_\theta (\mathbf{q}, \mathcal{P}) = \min_{\mathbf{p} \in \mathcal{P}} \hat{d}_{\theta}(\mathbf{q}, \mathbf{p})$, so that $h_\theta(\mathbf{q}, \mathcal{P}) \geq 0$ represents safety.
% \BL{Uncertainty in $h$ and $\mathcal{P}$}

To plan a trajectory that satisfies the NCSB constraint, we use the safe bubble cover~\cite{safe_bubble_cover} algorithm, which exploits the Lipschitz property of the NCSB $h_\theta$ to construct spherical safe regions around sampled configurations.
Doing so obviates collision checking within the safe regions, dramatically reducing computation time for faster re-planning. 
% \NA{Explain why this is (e.g., it is not necessary to perform collision checking within whole regions of space given by the safe bubbles, and explain how this relates to RRT, i.e., give an overview of the algorithm in 1-2 sentences.}.

Meanwhile, the plan does not consider uncertainties in the pointcloud $\mathcal{P}$ and the CDF parameters $\theta$.
To execute actions that are robustly safe against these uncertainties, we use the distributionally robust control barrier function (DR-CBF) formulation from our prior work~\cite{dro_cbf}:
\begin{equation}
\begin{aligned}
    &\mathbf{u}^\star = \min_{\mathbf{u}} || \mathbf{u} - \mathbf{u}_{nom} ||, \\
    \text{ s.t. } &\inf_{\mathbb{P} \in \mathcal{M}(\mathcal{P}, \theta)} \mathbb{P}\left( \frac{\partial h_\theta}{\partial \mathbf{q}}\mathbf{u} + \frac{\partial h_\theta}{\partial t} + \alpha h_\theta \geq 0 \right) \geq 1-\epsilon,
\end{aligned}
\end{equation}
where $\mathbf{u}_{\text{nom}}$ is a nominal control action from either the teleoperation device or the plan, and $\alpha \geq 0$ is a parameter. 
The probabilistic constraint ensures safety $h_\theta(\mathbf{q}, \mathcal{P}) \geq 0$ with probability $\geq 1 - \epsilon$ over the manipulator's trajectory, with respect to all distributions $\mathcal{M}(\mathcal{P}, \theta)$ over the pointcloud $\mathcal{P}$ and CDF parameters $\theta$ that are within a given Wasserstein distance away from the observations.
The constraint can be relaxed to a linear inequality over samples of $\mathcal{P}$ and $\theta$ ~\cite{dro_cbf}.

% \NA{Explain that $\mathcal{M}$ is a set of distributions that are within a certain (Wasserstein) distance from the sample distribution of point cloud and neural network parameter samples.}

%kmk
\subsection{Teleoperation}\label{sec:teleop}
\begin{figure}[!t]
    \centering
    \includegraphics[width=0.4\textwidth]{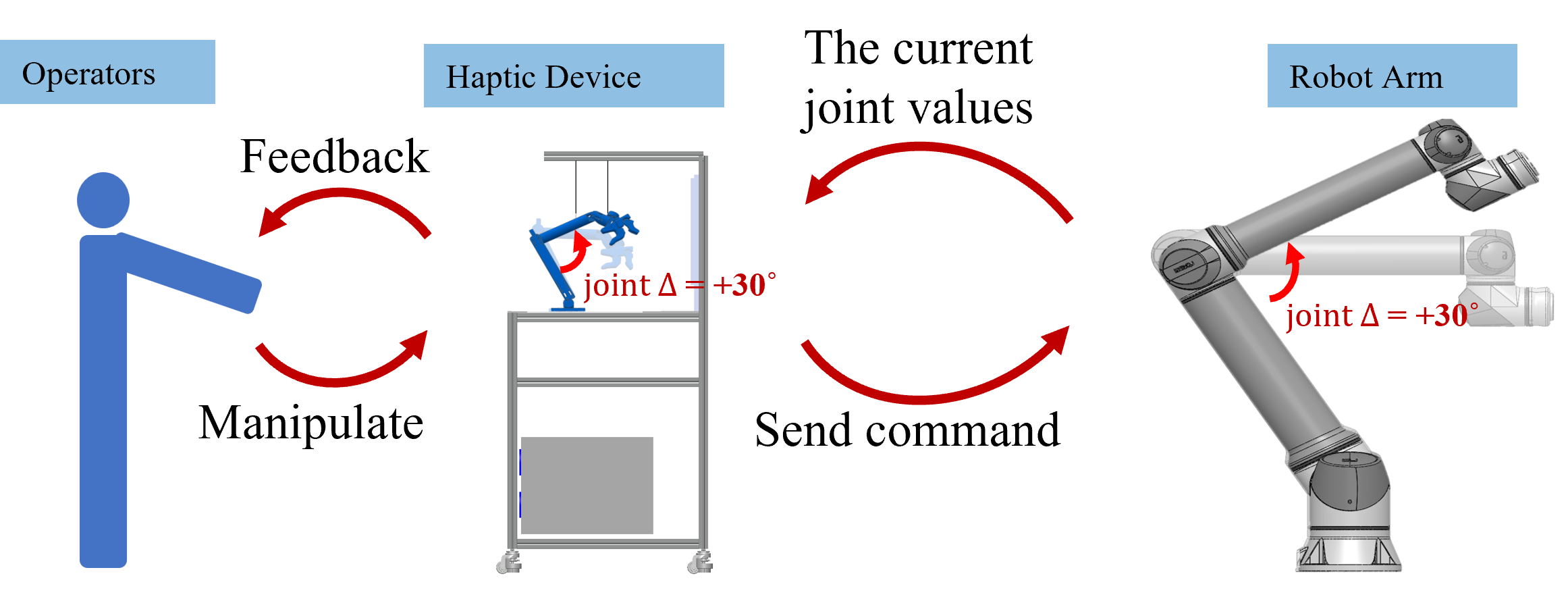}
    \caption{Flow diagram of teleoperation using the haptic device.}
    \label{fig:teleoperation}
\end{figure}

Teleoperating a manipulator can be counterintuitive due to the kinematic differences between humans and robots.
BVLOS teleoperation is even more challenging due to a lack of tactile feedback.
Thus, we build on GELLO~\cite{wu2023gello} to develop a haptic device that mirrors the kinematic structure of the manipulator, and provides proprioceptive feedback. We customized the GELLO hardware~\cite{wu2023gello} for the RB10-1300 manipulator, and developed 
a custom software illustrated in Fig.~\ref{fig:teleoperation}.
When the operator manipulates the haptic device, the software generates relative joint position commands for the manipulator by comparing the changes in the haptic device's joint angles to the current joint angles of the manipulator.
% The manipulator's joint positions are continuously read.
% Using a continuous stream of the manipulator joint values, 
% The current joint values of the manipulator are continuously received and stored. 
% When the haptic device is manipulated, it calculates the changes in joint angles and applies them to the stored values to generate real-time control commands. 
This allows the operator to fully control all six joints of the manipulator.
For proprioceptive feedback, actuator currents are applied to the haptic device that are proportional to those measured in the manipulator, replicating the corresponding torques.
Thus, the device provides proprioceptive feedback that emulates the actual physical interaction, enhancing realism and intuition for effective operation. 

% In BVLOS settings, the operator often lack tactile realism. To address this issue, we additionally measure the current flowing through each joint of the RB10-1300 and apply proportional current feedback to the corresponding joints of the haptic device. 
% This approach provides tactile feedback that simulates actual physical interaction, thereby enhancing realism and allowing for more intuitive and effective operation.

 %kmk

\section{Feasibility Study}
\subsection{Robot Drilling}
\begin{figure}[!t]
    \noindent
    \subcaptionbox{Three different experimental sites.\label{fig:environment_test_sites}}
    {
    \includegraphics[width=0.15\textwidth]{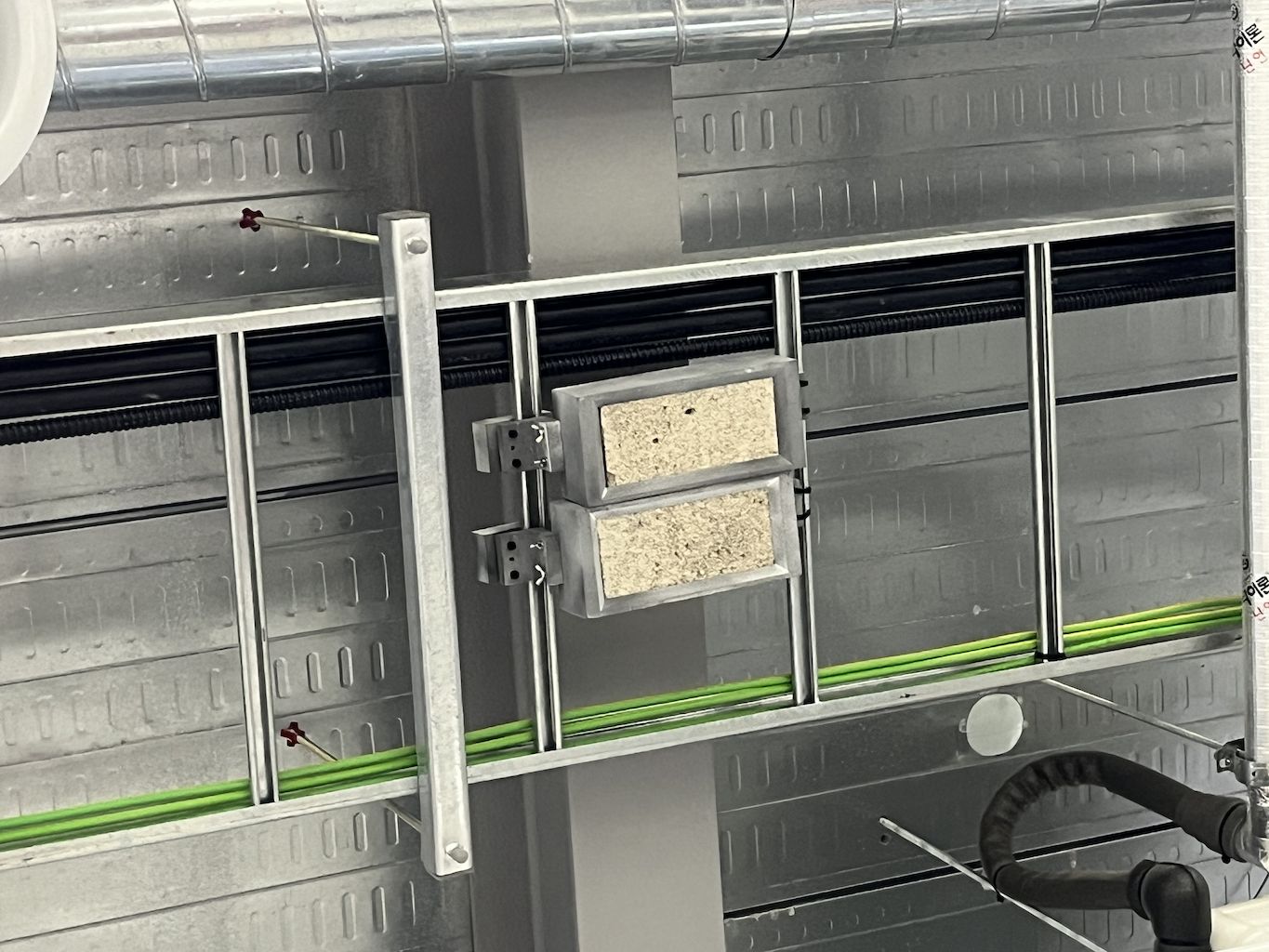}\hspace{2mm}
    \includegraphics[width=0.15\textwidth]{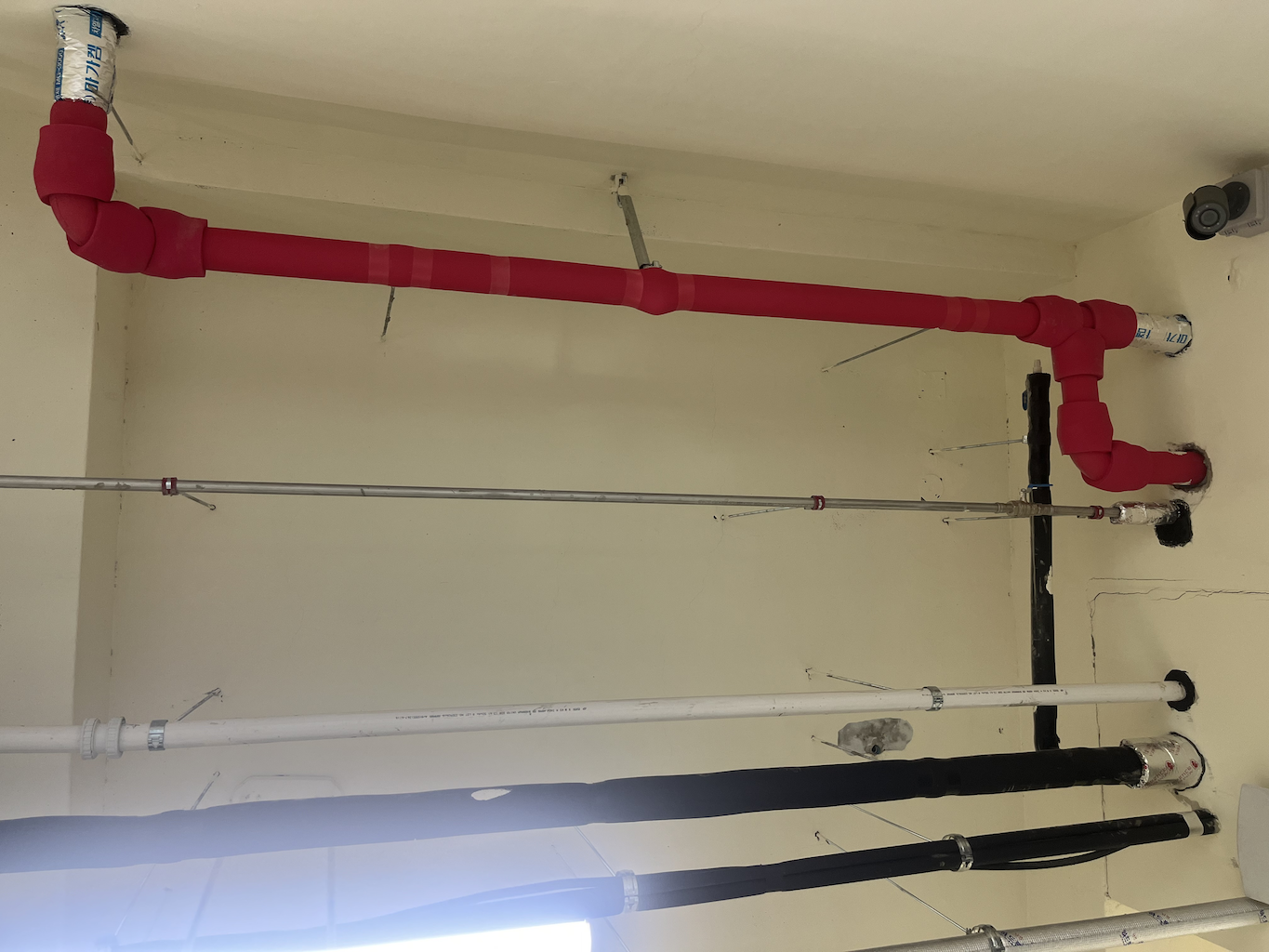}\hspace{2mm}
    \includegraphics[width=0.15\textwidth]{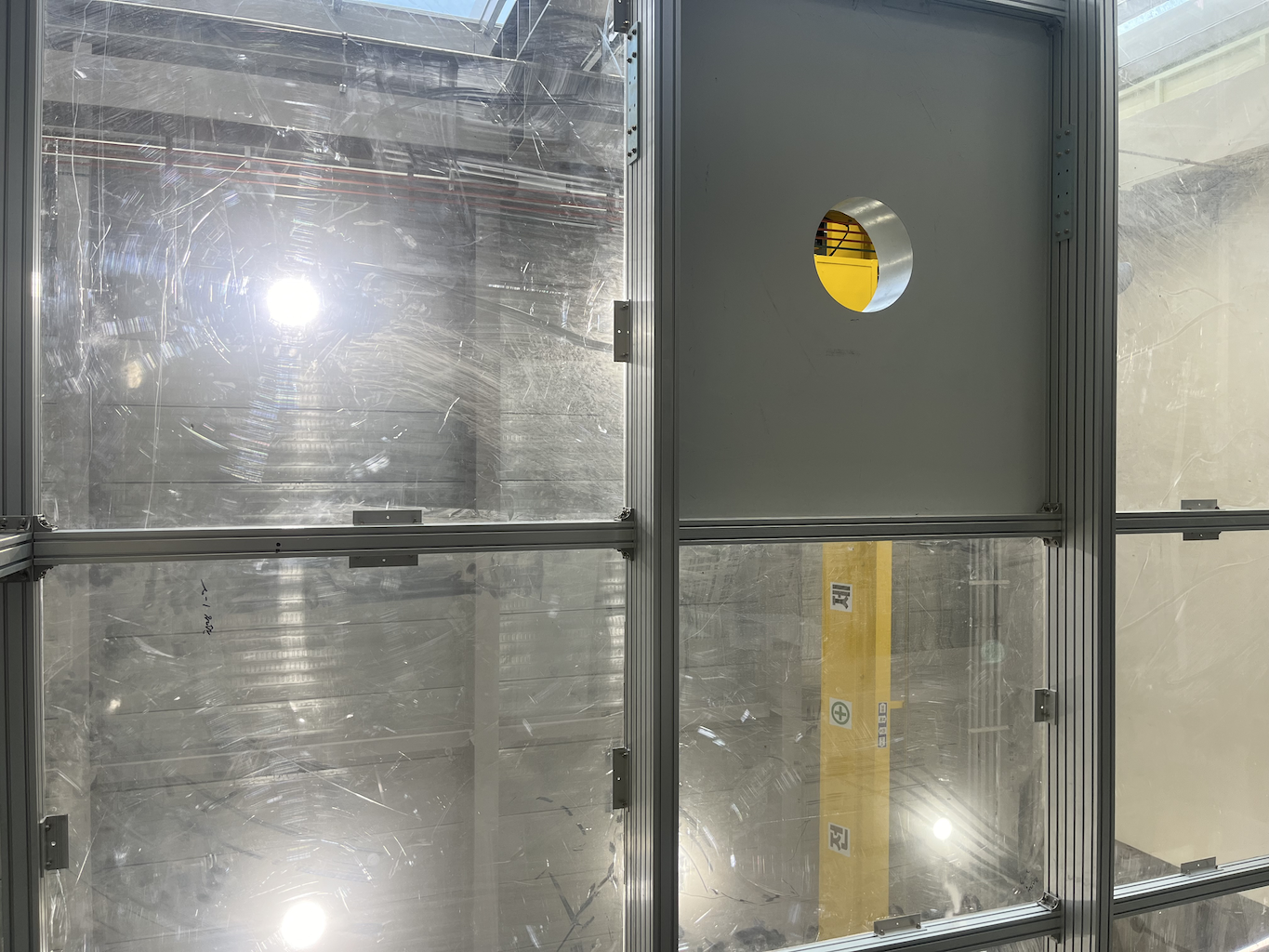}
    }
    % \vspace{-2ex}  % 여백 줄이기
    % \begin{center}
        % \subcaption*{(a) Three different experimental environments.}
    % \end{center}
    % \noindent
    \subcaptionbox{Results from each site}
    {
    \includegraphics[width=0.15\textwidth]{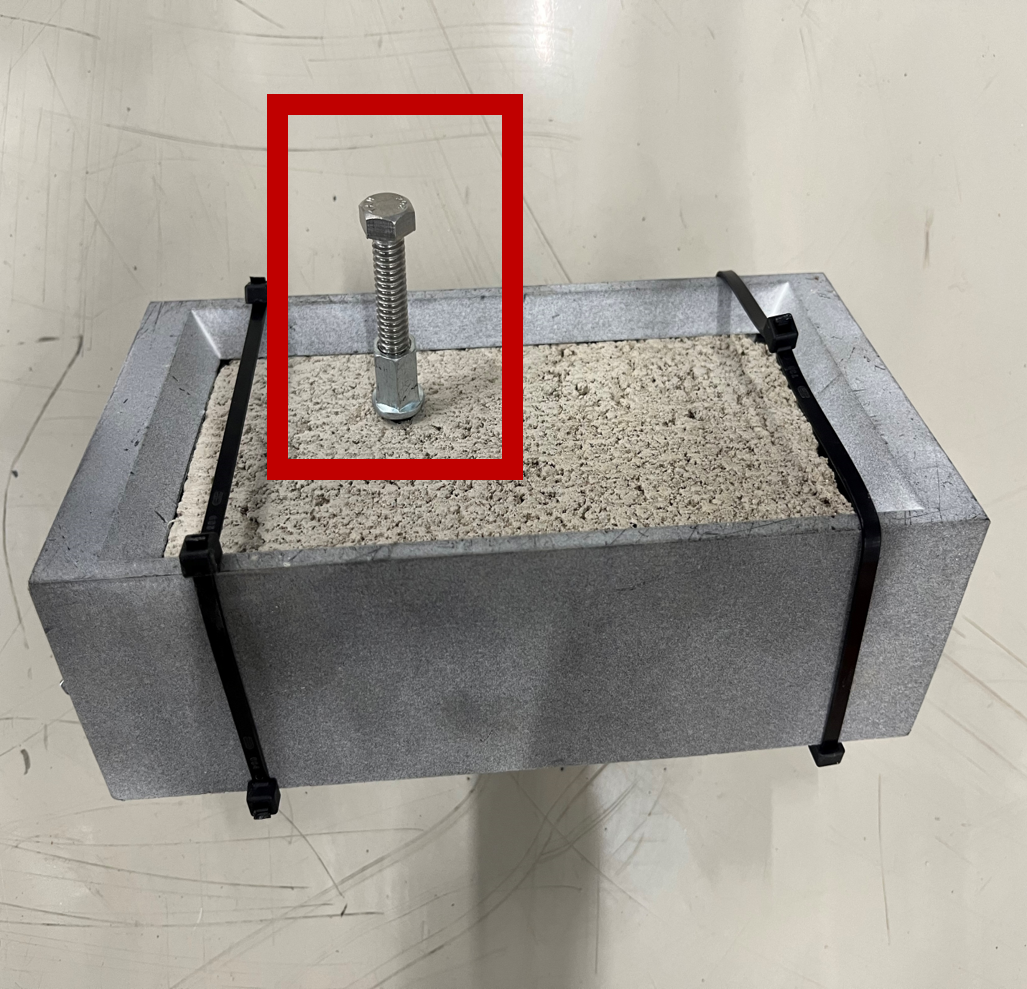}\hspace{2mm}
    \includegraphics[width=0.15\textwidth]{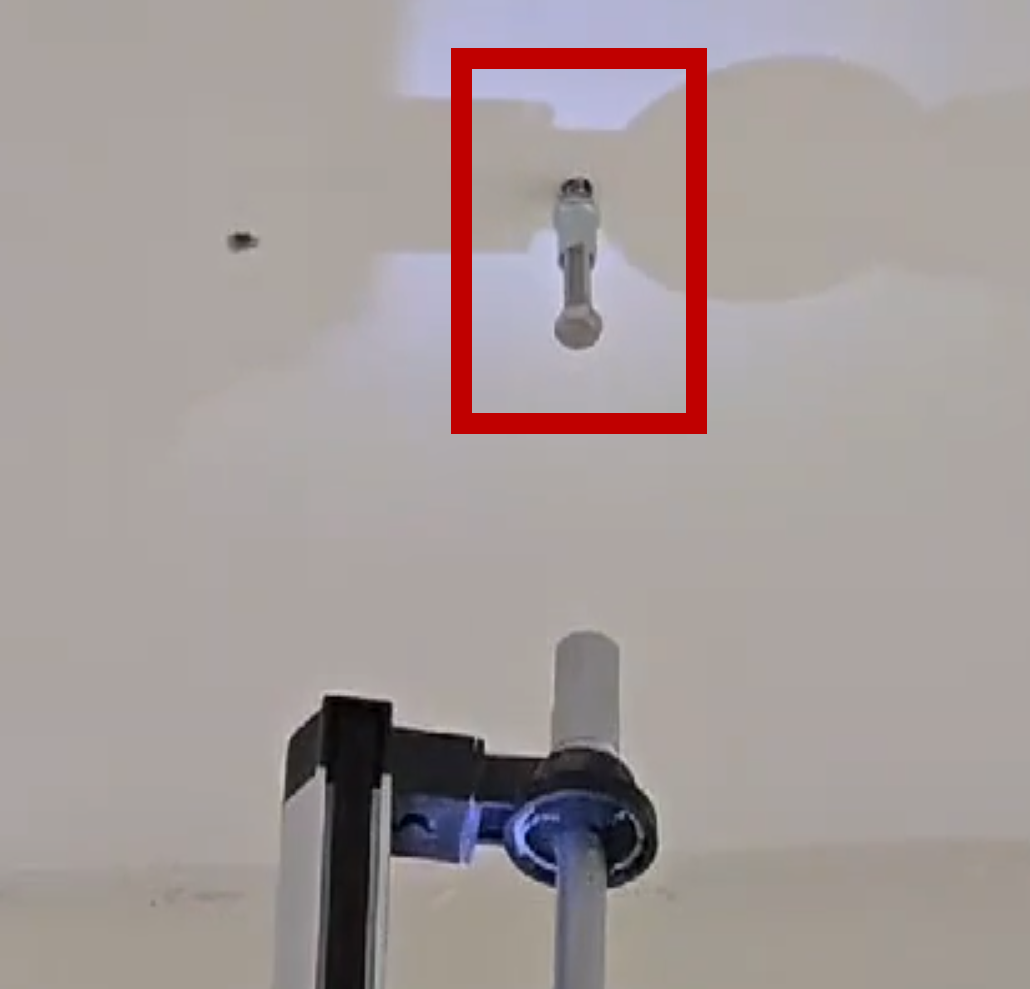}\hspace{2mm}
    \includegraphics[width=0.15\textwidth]{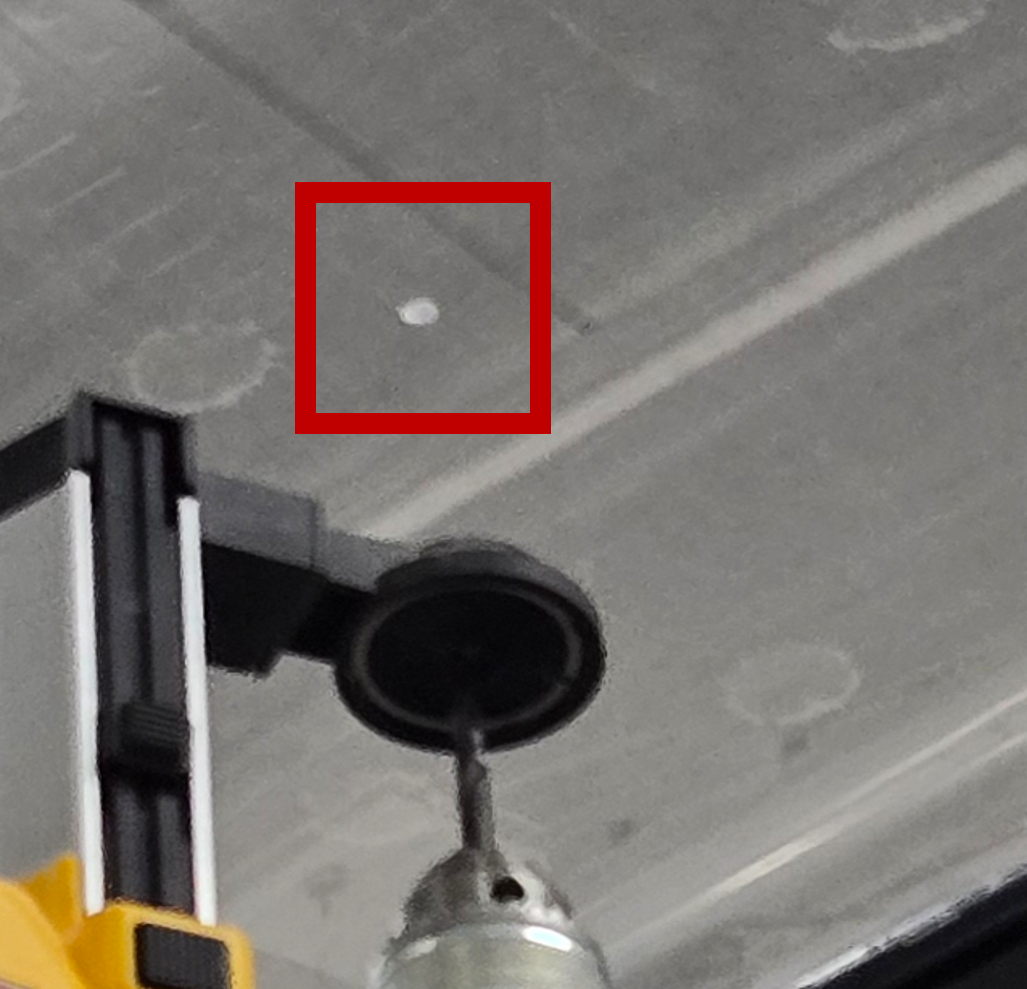}
    }
    % \vspace{-2ex}  % 여백 줄이기
    % \begin{center}
        % \subcaption*{(b) Results from each environment}
    %     \vspace{-2ex}  % 여백 줄이기
    \caption{Experimental sites and results.}\label{fig:environment_test}
    % \end{center}   
\end{figure}

We evaluate the feasibility of the hardware platform for BVLOS teleoperation by performing drilling, anchoring, and bolting on ceilings, which are representative of overhead works. 
We considered ceilings made of concrete, gypsum, and acrylic materials in three different sites, as shown in Fig.~\ref{fig:environment_test}(a), for a total of 9 experiments.

\begin{table}[h]
\centering
\caption{Result of the experiments conducted with three different material types in all sites in Fig.~\ref{fig:environment_test}(a)}
\label{tab:experiment_env1}
\renewcommand{\arraystretch}{1.1} % Row height adjustment
\setlength{\tabcolsep}{3pt} % Column spacing adjustment
\begin{tabular}{c|c|c|c}
\hline
\multirow{2}{*}{Method} & \multicolumn{3}{c}{Experimental Materials} \\
\cline{2-4}
 & Concrete & Gypsum & Acrylic \\
\hline
Drilling & \checkmark & \checkmark & \checkmark \\
Anchoring & \checkmark & \checkmark & X \\
Bolting & \checkmark & \checkmark & X \\
\hline
\end{tabular}
\end{table}

\begin{figure*}[h]
    \centering
    % First row
    \subcaptionbox{Planned configurations\label{fig:6a}}
    {\includegraphics[width=0.24\textwidth]{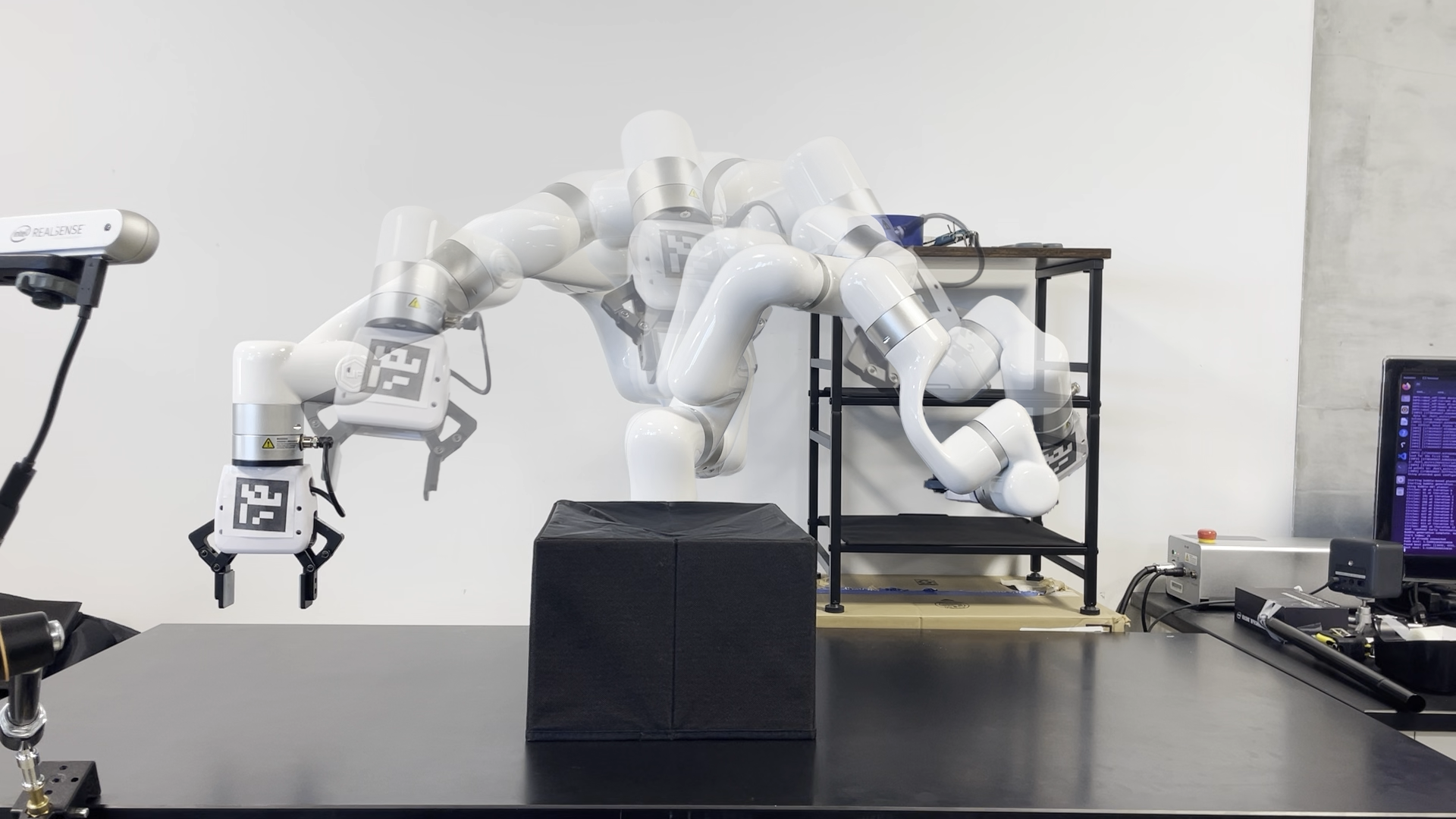}}
    \subcaptionbox{Response to dynamic obstacle\label{fig:6b}}
    {\includegraphics[width=0.24\textwidth]{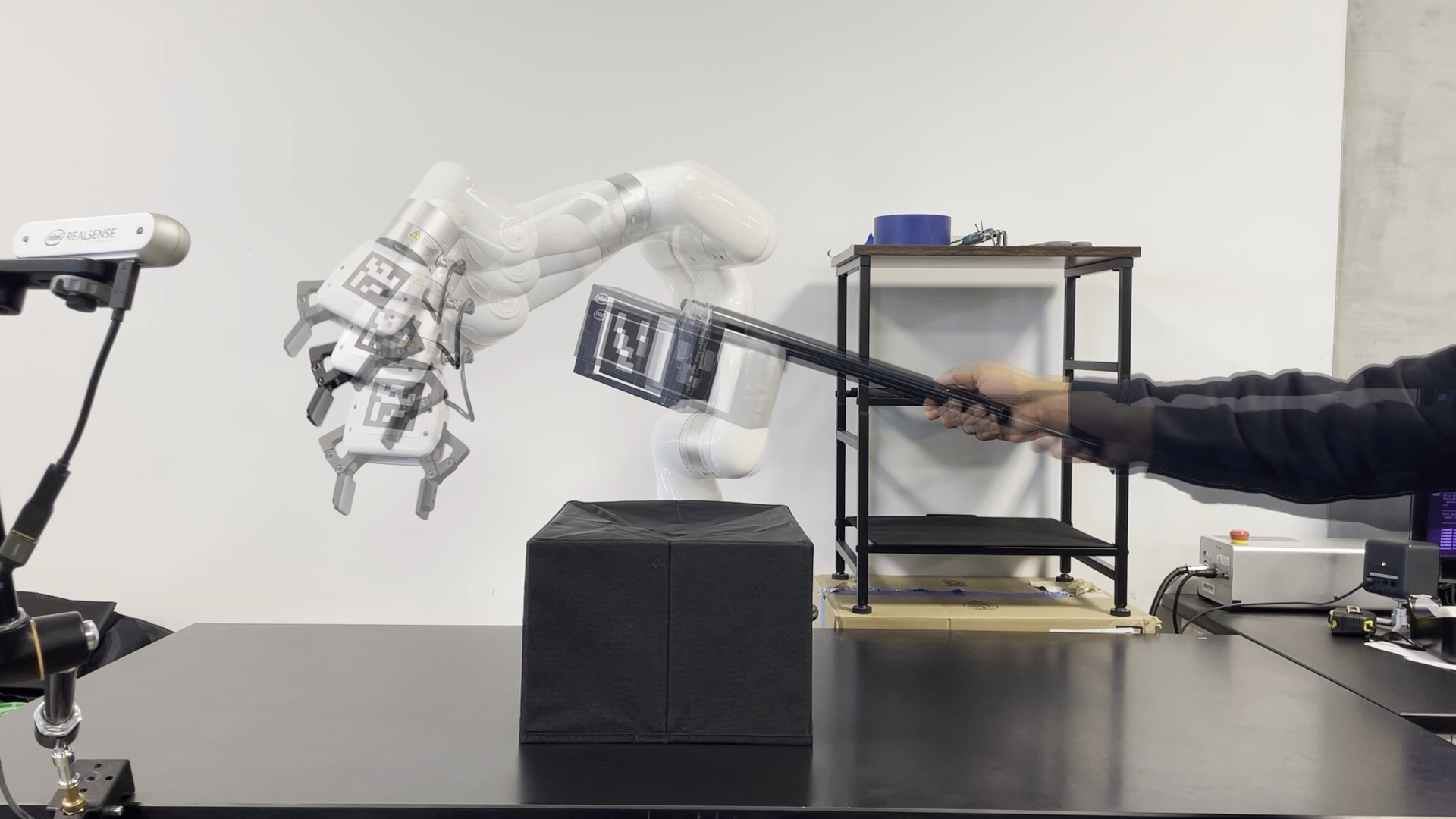}}
    \subcaptionbox{Response to dynamic obstacle\label{fig:6c}}
    {\includegraphics[width=0.24\textwidth]{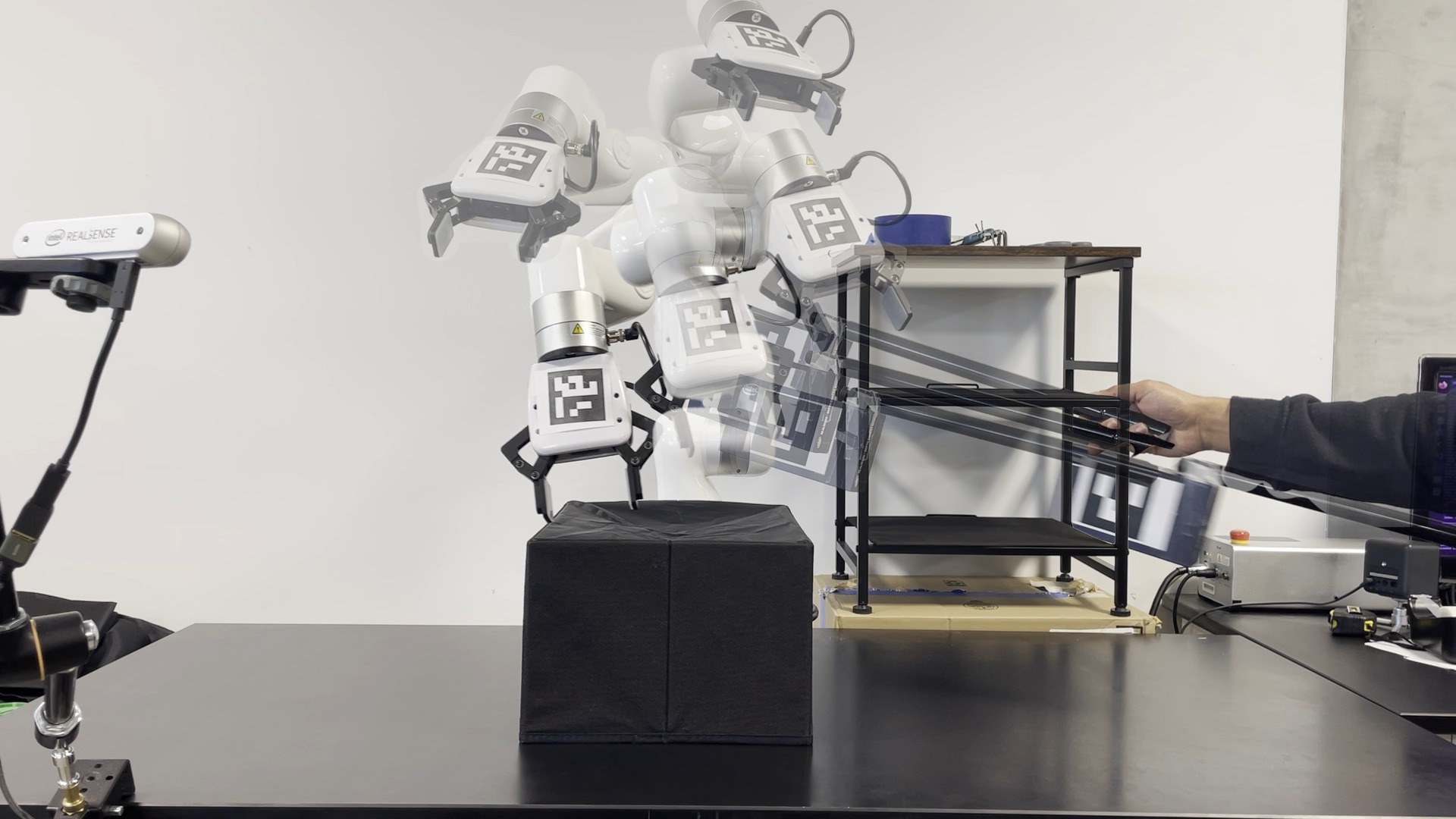}}
    \subcaptionbox{Response to dynamic obstacle\label{fig:6d}}
    {\includegraphics[width=0.24\textwidth]{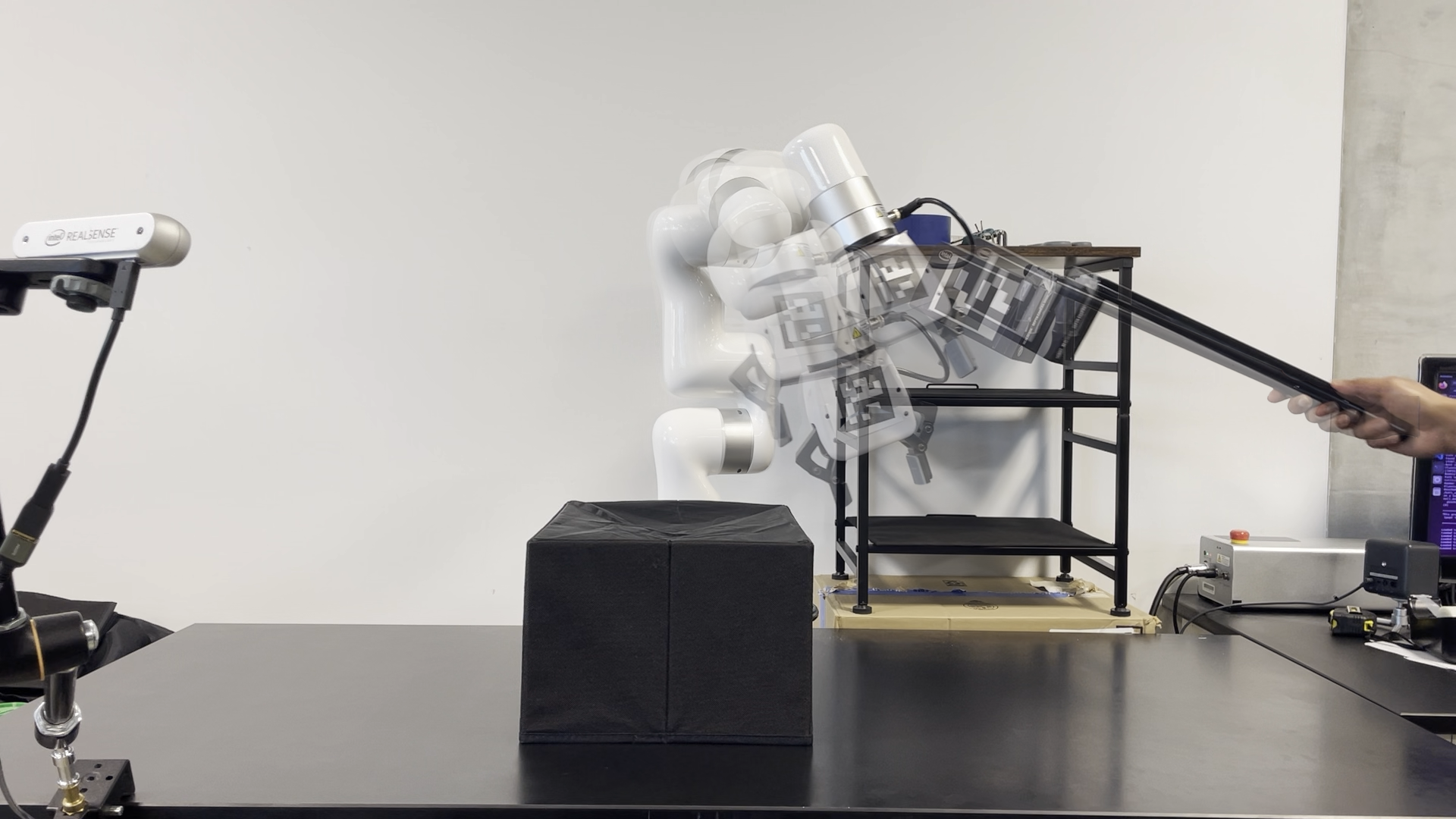}} \\
    % \subcaptionbox{Planned configurations\label{fig:7a}}
    % {\includegraphics[width=0.24\textwidth]{ieeeconf/3.Feasibility_Study/figure/Top_Two_Box_Composites/top_two_box_static.png}}
    % \subcaptionbox{Response to dynamic obstacle \label{fig:7b}}
    % {\includegraphics[width=0.24\textwidth]{ieeeconf/3.Feasibility_Study/figure/Top_Two_Box_Composites/top_two_box_dynamic_1.png}}
    % \subcaptionbox{Response to dynamic obstacle\label{fig:7c}}
    % {\includegraphics[width=0.24\textwidth]{ieeeconf/3.Feasibility_Study/figure/Top_Two_Box_Composites/top_two_box_dynamic_2.png}}
    % \subcaptionbox{Response to dynamic obstacle\label{fig:7d}}
    % {\includegraphics[width=0.24\textwidth]{ieeeconf/3.Feasibility_Study/figure/Top_Two_Box_Composites/top_two_box_dynamic_3.png}}  
    \caption{Bubble-CDF planner and DR-CBF control in a laboratory experiment with a 6-DoF xArm robot. }
    \label{fig: real_world_demo}
    \vspace*{-1ex}
\end{figure*}

As shown in Fig.~\ref{fig:environment_test}(b), all tasks were successfully performed on concrete and gypsum materials in all three sites. For acrylic, only drilling succeeded, whereas anchoring and bolting failed in all sites. The failure of the two operations is because they rely on frictional forces, which acrylic does not provide due to its smoothness. In contrast, the rougher surfaces of concrete and gypsum offered enough friction for stable anchoring and bolting.
We are exploring bimanual operation to address this limitation. 
% In future work, we aim to investigate this approach further to improve task success rates on low-friction materials such as acrylic.

\subsection{Planning and Control}
We present an initial evaluation of the suitability of the planning and control modules in dynamic environments using a 6-DoF xArm6~\cite{xarm6} in a laboratory environment.
The manipulator was assigned a goal end-effector position to navigate towards in a cluttered environment with both static and dynamic obstacles as depicted in Fig. ~\ref{fig: real_world_demo}.
% As described in~\cite{neural_cspace_barrier}, the safe bubble cover~\cite{safe_bubble_cover} algorithm is used for generating a nominal path, while the DR-CBF~\cite{dro_cbf} is used to avoid any dynamic obstacles that were unaccounted for.
A single Realsense D-435 camera was used to provide observations of the obstacles from a fixed location.
We used an Aruco marker to track the velocity of dynamic obstacles, which may be replaced with DynaGSLAM.

% We conducted further experiments to evaluate the planning and control architecture by having a 6-DoF xArm navigate in the cluttered environment to reach a specific configuration  The bubble-CDF planner was creating a nominal trajectory to follow by constructing safe-bubbles in the configuration space, which represented the collision-free arm configurations, and based on that, the trajectory was generated in this free space to reach a desired arm configuration. 
% The arm configuration was usually specified by the end effector between the two shelves (e.g,. top or bottom shelf). Since the environment was dynamic with some obstacles moving in space, similar to unpredictable environments in construction sites, the DR-CBF controller was used, which takes as input the environment point clouds from the depth camera and, based on their position and velocity avoids the incoming obstacle if it's within the safety radius. 

The results in Fig.~\ref{fig: real_world_demo} show that the manipulator follows a nominal trajectory to reach the desired goal configuration, while deviating from the nominal trajectory to avoid incoming obstacles as necessary. 
% , at which point it does not follow the nominal policy anymore but tries only to avoid the moving object. 
% The experiments were repeated in several cluttered and dynamic environments with different goal configurations for a total of 20 randomized trials with varying obstacle placements.
We observed that the safe bubble cover~\cite{safe_bubble_cover} significantly reduces the number of collision checks, by up to tenfold compared to other baselines, while yielding comparable path lengths.
% The main observations from the experiments were that the bubble-CDF planner significantly reduces the number of collisions checks compared to all baselines in the magnitude of a tenfold collision reduction because it uses configuration space distance to obtain a large space of collision-free regions. 
% The path length remained comparable to other baselines, indicating convergence to the optimum even with reduced computation.

\begin{figure}
    \centering
    \subcaptionbox{Construction Dataset~(Fig.~\ref{fig:environment_test})}
    {
    \includegraphics[width=0.3\linewidth]{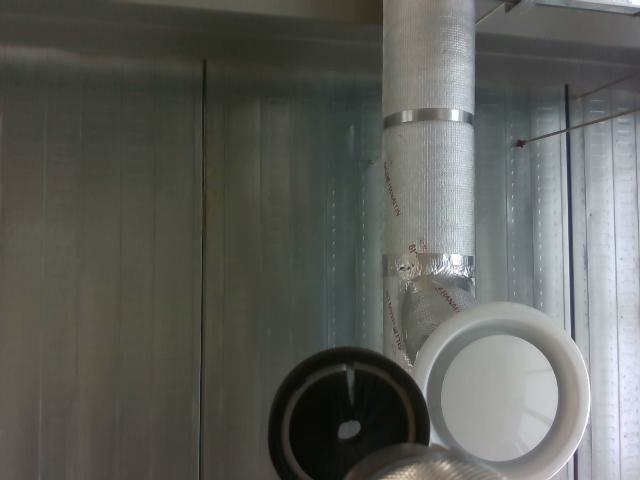}
    \includegraphics[width=0.3\linewidth]{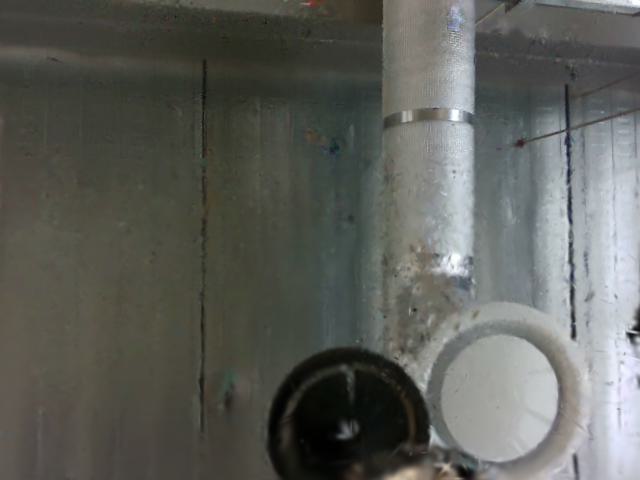}
    \includegraphics[width=0.3\linewidth]{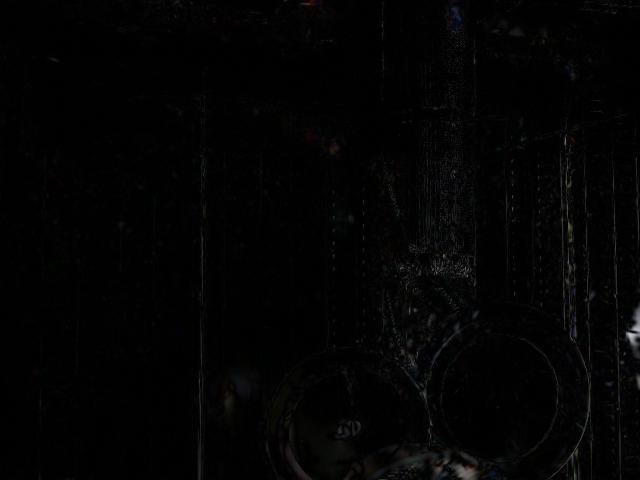}    
    }
    \subcaptionbox{TUM Dataset~\cite{tum}}
    {
    \includegraphics[width=0.3\linewidth]{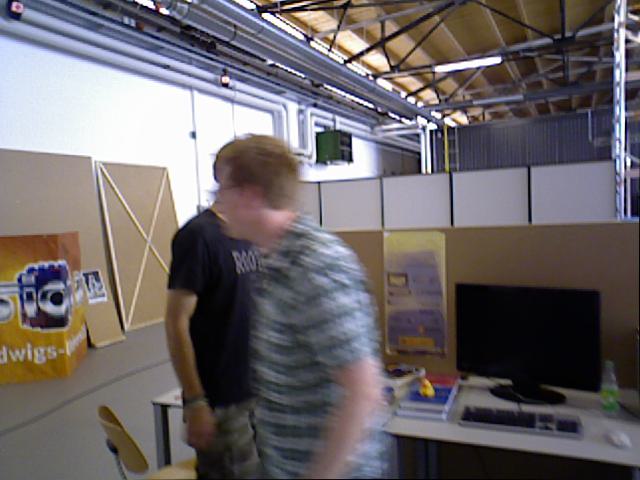}
    \includegraphics[width=0.3\linewidth]{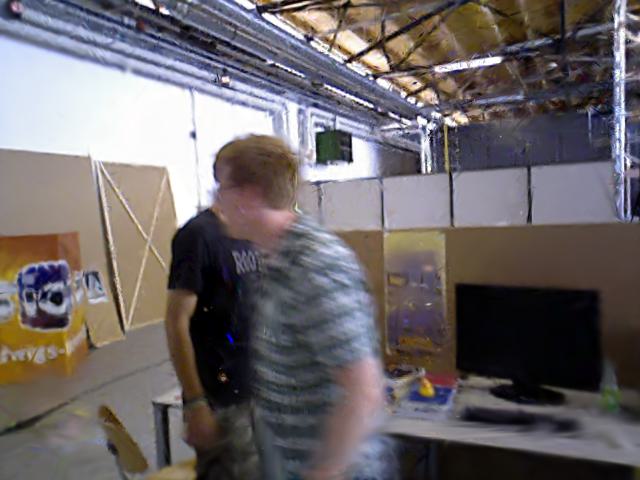}
    \includegraphics[width=0.3\linewidth]{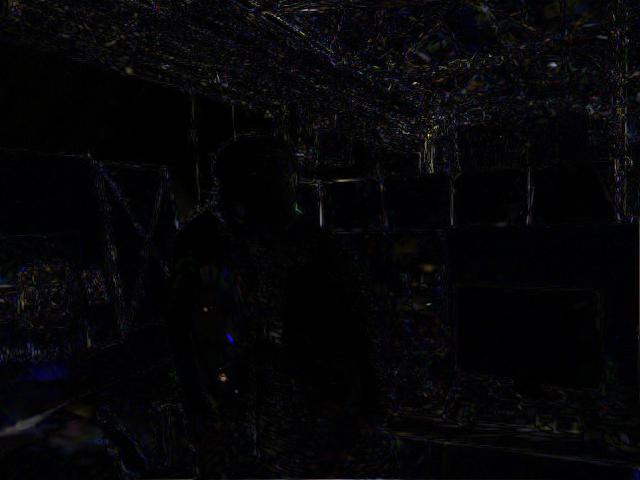}
    }    
    \caption{Qualitative results of DynaGSLAM. Left: ground truth. Center: rendered. Right: error.}
    \label{fig:dynagslam}
\end{figure}

\begin{table}[h]
\centering
\caption{Image Reconstruction Accuracy of DynaGSLAM}
\label{tab:experiment_dynagslam}
% \renewcommand{\arraystretch}{1.1} % Row height adjustment
% \setlength{\tabcolsep}{3pt} % Column spacing adjustment
% \begin{tabular}{c|c|c|c}
% \hline
% \multirow{2}{*}{Method} & \multicolumn{3}{c}{Experimental Materials} \\
% \cline{2-4}
%  & Concrete & Gypsum & Acrylic \\
% \hline
% Drilling & \checkmark & \checkmark & \checkmark \\
% Anchoring & \checkmark & \checkmark & X \\
% Bolting & \checkmark & \checkmark & X \\
% \hline
% \end{tabular}
\begin{tabular}{c|c|c|c}
Dataset & PSNR (dB)$\uparrow$ & LPIPS (\%)$\downarrow$ & SSIM (\%)$\uparrow$  \\
               \hline
TUM~\cite{tum} & 27.2           &   20.0 & 94.5  \\
               \hline
Construction (Fig.~\ref{fig:environment_test}) & 26.8   &   44.7 & 84.4 
\end{tabular}

\end{table}

\subsection{Dynamic Mapping}
We evaluate dynamic scene reconstruction with DynaGSLAM in two datasets, TUM~\cite{tum} and one collected during ceiling drilling experiments (Fig.~\ref{fig:environment_test}).
The qualitative results in Fig.~\ref{fig:dynagslam} and the quantitative results in Table~\ref{tab:experiment_dynagslam} show that DynaGSLAM renders RGB images accurately.
This is owing to accurate reconstruction of dynamic environments despite moving objects and featureless ceilings.
Such ability of handling moving objects allows photorealistic visual feedback even as the manipulator or work pieces are moving.
% The GS are optimized online in real time in the SLAM system, which facilitates the rendering of robot-object interactions from novel unseen camera views. 
% It is important to render moving objects in the scene for robotic manipulation, as the end effectors and objects are moving during manipulations.
% We build our DynaGSLAM to support the rendering of dynamic entities in the scene, and we tested the dynamic rendering in TUM and Bonn datasets. 
% As the first GS-based dynamic SLAM work, our DynaGLSAM outperforms all the baseline GS-SLAM that could only handle static scenes. %kmk

\section{Ongoing and Future Work}
\begin{figure}[!t]
    % \centering
    % First row
    \subcaptionbox{Dual-arm construction robot}
    {\includegraphics[width=0.58\linewidth]{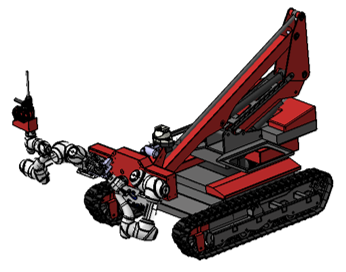}}%
    \hfill%
    \subcaptionbox{Teleoperation booth}
    {\includegraphics[width=0.42\linewidth]{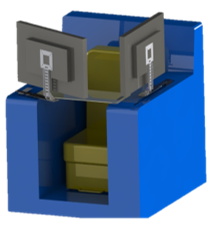}}
    \caption{Envisioned hardware improvements}
    \label{fig: future_work}
    \vspace{-2ex}
\end{figure}

% Future and Ongoing Work
We presented an initial hardware prototype and the experimental results of individual software modules in isolation in laboratory settings.
We are currently upgrading the hardware platform based on initial findings, as envisioned in Fig.~\ref{fig: future_work}, and are integrating the software modules for a full demonstration in a real construction site.
Specifically, we will mount an additional manipulator to the torso, which will enable simultaneous handling of fasteners and fixtures used in ceiling installations.
The planning and control module will be accordingly extended to jointly consider bimanual operations and the mobile base.
We are also integrating higher-payload manipulators to support more demanding tasks, and a more permanent booth will be built for greater ergonomic comfort.
% In addition, to support safer and more complex tasks, the RB10-1300 manipulator will be replaced with a higher payload manipulator in the future.

We are also making further improvements to the guidance software for greater autonomy.
In particular, the teleoperation interface will allow the operator to specify a goal end-effector pose for planning.
To support planning, we are working on constructing the NCSB from Gaussian splat maps. 
% The Gaussian splat map 

% Ver 2
% \subsection{Future Improvements in Manipulator Design}
% 양팔 로봇 장착 예정
% payload가 더 좋은 로봇팔로 교체 예정

% refer draft,  syh
\bibliographystyle{IEEEtran}
\bibliography{ref}

% Generated by IEEEtran.bst, version: 1.14 (2015/08/26)
\begin{thebibliography}{10}
\providecommand{\url}[1]{#1}
\csname url@samestyle\endcsname
\providecommand{\newblock}{\relax}
\providecommand{\bibinfo}[2]{#2}
\providecommand{\BIBentrySTDinterwordspacing}{\spaceskip=0pt\relax}
\providecommand{\BIBentryALTinterwordstretchfactor}{4}
\providecommand{\BIBentryALTinterwordspacing}{\spaceskip=\fontdimen2\font plus
\BIBentryALTinterwordstretchfactor\fontdimen3\font minus
  \fontdimen4\font\relax}
\providecommand{\BIBforeignlanguage}[2]{{%
\expandafter\ifx\csname l@#1\endcsname\relax
\typeout{** WARNING: IEEEtran.bst: No hyphenation pattern has been}%
\typeout{** loaded for the language `#1'. Using the pattern for}%
\typeout{** the default language instead.}%
\else
\language=\csname l@#1\endcsname
\fi
#2}}
\providecommand{\BIBdecl}{\relax}
\BIBdecl

\bibitem{constructiondive2018}
K.~S. Anderson, ``Survey: 65\% of construction workers are fatigued on the
  job,'' \emph{Construction Dive}, 2018.

\bibitem{Schwatka2012}
N.~V. Schwatka, L.~M. Butler, and J.~R. Rosecrance, ``An aging workforce and
  injury in the construction industry,'' \emph{Epidemiologic Reviews}, vol.~34,
  pp. 156--167, 2012.

\bibitem{hilti_jaibot}
\BIBentryALTinterwordspacing
{Hilti Group}, ``Hilti jaibot – semi-autonomous drilling robot for ceiling
  work,'' 2025, accessed: 2025-04-17. [Online]. Available:
  \url{https://www.hilti.com/content/hilti/W1/US/en/business/business/trends/jaibot.html}
\BIBentrySTDinterwordspacing

\bibitem{csc_drillcorpio}
\BIBentryALTinterwordspacing
{CSC Robotic Engineering}, ``Drillcorpio – ceiling construction robot,''
  2025, accessed: 2025-04-17. [Online]. Available:
  \url{https://cscrobotic.com/en/our-products/}
\BIBentrySTDinterwordspacing

\bibitem{dynagslam}
R.~B. Li, M.~Shaghaghi, K.~Suzuki, X.~Liu, V.~Moparthi, B.~Du, W.~Curtis,
  M.~Renschler, K.~M.~B. Lee, N.~Atanasov, and T.~Nguyen, ``Dyna{GSLAM}:
  Real-time {G}aussian-splatting {SLAM} for online rendering, tracking, motion
  predictions of moving objects in dynamic scenes,'' \emph{arXiv preprint
  arXiv:2503.11979}, 2025.

\bibitem{drones2020013}
S.~X. Fang, S.~O’Young, and L.~Rolland, ``Development of small {UAS}
  beyond-visual-line-of-sight flight operations: System requirements and
  procedures,'' \emph{Drones}, vol.~2, no.~2, 2018.

\bibitem{neural_cspace_barrier}
K.~Long, K.~M.~B. Lee, N.~Raicevic, N.~Attasseri, M.~Leok, and N.~Atanasov,
  ``Neural configuration-space barriers for manipulation planning and
  control,'' \emph{arXiv preprint arXiv:2503.04929}, 2025.

\bibitem{Tiberkak2023WebRTC}
A.~Tiberkak, A.~Hentout, and A.~Belkhir, ``Web{RTC}-based {MOSR} remote control
  of mobile manipulators,'' \emph{International journal of intelligent robotics
  and applications}, vol.~7, no.~2, p. 304—320, 2023.

\bibitem{wu2023gello}
P.~Wu, Y.~Shentu, Z.~Yi, X.~Lin, and P.~Abbeel, ``Gello: A general, low-cost,
  and intuitive teleoperation framework for robot manipulators,'' 2023, arXiv
  preprint arXiv:2307.01234.

\bibitem{li2024cdf_rss}
Y.~Li, X.~Chi, A.~Razmjoo, and S.~Calinon, ``Configuration space distance
  fields for manipulation planning,'' in \emph{Robotics: Science and Systems
  (RSS)}, 2024.

\bibitem{safe_bubble_cover}
K.~M.~B. Lee, Z.~Dai, C.~L. Gentil, L.~Wu, N.~Atanasov, and T.~Vidal-Calleja,
  ``Safe bubble cover for motion planning on distance fields,'' \emph{arXiv
  preprint arXiv:2408.13377}, 2024.

\bibitem{dro_cbf}
K.~Long, Y.~Yi, Z.~Dai, S.~Herbert, J.~Cortés, and N.~Atanasov, ``Sensor-based
  distributionally robust control for safe robot navigation in dynamic
  environments,'' \emph{arXiv preprint arXiv:2405.18251}, 2024.

\bibitem{xarm6}
\BIBentryALTinterwordspacing
{UFactory}, ``{xArm},'' 2025, accessed: 2025-04-17. [Online]. Available:
  \url{https://www.ufactory.us/xarm}
\BIBentrySTDinterwordspacing

\bibitem{tum}
J.~Sturm, N.~Engelhard, F.~Endres, W.~Burgard, and D.~Cremers, ``A benchmark
  for the evaluation of {RGB-D SLAM} systems,'' in \emph{Proc. of the
  International Conference on Intelligent Robot Systems (IROS)}, Oct. 2012.

\end{thebibliography}
% Intro

% - Worker may be remote / not on-site

% Hardware Platform

% - Lift + Manipulator
%     - Ver. 1, RB-10 ⇒ Ver. 2 Doosan
%     - Torque specification
%     - Mounting design
%     - End-effector
%     - Sensors
% - Mobile Base
%     - Hardware is off-the-shelf, software
%     - TCP control software
% - Teloperation Booth
%     - Design figure
%     - Haptic Sensing
% - Specification Table (IT-ONE)

% Software

% - Block Diagram
%     - D455 ⇒ SDF Mapping ⇒ Planning ⇒ CBF Safety Filter
%     - Teleoperation ⇒ CBF Safety Filter
% - Autonomy
%     - Bubble planning + CBF safety filter
%     - Octomap
%     - SDF
% - Teleoperation + Shared Autonomy
%     - Anomaly sensing, reporting
%     - Haptic sensing

% Feasibility Study

% - Robot Drilling, comparison of teleop/shared/feedforward
% - Collision avoidance (CBF)

% Future and Ongoing Work

% - Integration of planning
% - Occlusion handling - Gaussian splatting, occlusion prediction
% - Ver. 2

\end{document}